# Testing the 'Efficient Network TRaining (ENTR)' Hypothesis: initially reducing training image size makes Convolutional Neural Network training for image recognition tasks more efficient

Thomas Cherico Wanger[1$*], Peter Frohn[2$]

**Affiliations:**
[1] Department of Crop Sciences, University of Göttingen, Germany
[2] Department of Mechanical Engineering, University of Siegen, Germany

*****Author for correspondence (TCW):**
Email: tomcwanger@gmail.com

**$ Author Contributions:**
TCW conceived the idea, sourced and compiled the Bees dataset, analysed the data and wrote the paper. PF sourced and compiled the Steel dataset and wrote the dataset description in the paper. TCW and PF discussed and contributed to different versions of the manuscript.

**Abstract**

Convolutional Neural Networks (CNN) for image recognition tasks are seeing rapid advances in the available architectures and how networks are trained based on large computational infrastructure and standard datasets with millions of images. In contrast, performance and time constraints for example, of small devices and free cloud GPUs necessitate efficient network training (i.e., highest accuracy in the shortest inference time possible), often on small datasets. Here, we hypothesize that initially decreasing image size during training makes the training process more efficient, because pre-shaping weights with small images and later utilizing these weights with larger images reduces initial network parameters and total inference time. We test this 'Efficient Network TRaining (ENTR) Hypothesis' by training pre-trained Residual Network (ResNet) models (ResNet18, 34, & 50) on three small datasets (steel microstructures, bee images, and geographic aerial images) with a free cloud GPU. Based on three training regimes of i) not, ii) gradually or iii) in one step increasing image size over the training process, we show that initially reducing image size increases training efficiency consistently across datasets and networks. We interpret these results mechanistically in the framework of regularization theory. Support for the ENTR hypothesis is an important contribution, because network efficiency improvements for image recognition tasks are needed for practical applications. In the future, it will be exciting to see how the ENTR hypothesis holds for large standard datasets like ImageNet or CIFAR, to better understand the underlying mechanisms, and how these results compare to other fields such as 'structural learning'.

**Introduction**

Since Alex Krizhevsky et al.[1] published their landmark paper on Convolutional Neural Networks (CNN) in 2012, the field has seen tremendous advances in the available architectures and the way how data is analysed. A CNN is built on convolution, polling and non-linear layers, with different ways to improve computational performance and reduce overfitting, for instance by regularization techniques. After the AlexNet[1], VGG[2], and Inception models[3] using different network depths, kernel sizes of the convolutions, and batch normalization, Residual Network (ResNet) models have taken over the image recognition domain. ResNet models use 'identity shortcut connections' that skip one or several weighted layers[4] and these so called 'residual blocks' are then stacked together. This network structure overcomes common problems of deep networks such as a 'vanishing gradient' (i.e., the gradient is back-propagated to earlier layers, making it infinitively small, and as the network goes deeper, its performance stagnates)[5], they are easy to train (for a review see[6]), and reach fast convergence[7].

Previous studies have compared the above architectures and design modifications in terms of non-linearity, pooling variants, network width, image pre-processing, and learning parameters, amongst others[8,9]. Usually both, work on architectures and technological advances are benchmarked on standard datasets with several tens of thousands to more than a million images such as ImageNet or CIFAR[6,10,11]. While this approach makes image recognition research comparable, some of the cutting edge approaches require access to significant hardware (e.g., 'AlphaGo'[12]) that is – in the case of AI power houses – currently worth several million US Dollars and it still takes weeks to train sophisticated models. However, there is increasing awareness of AI across scientific disciplines and industry domains, and the availability of open-source tools allows practitioners to work on relevant problems, often with limited computing power and relatively small datasets[13]. Moreover, increasingly small devices rely on less performance hungry approaches to train networks. Thus, the resource gradient and increasing demand for low performance devices requires procedures to train state-of-the-art architectures like ResNet on small datasets more efficiently.

Performance improvements of convolutional neural networks involve accuracy, inference time, and the number of estimated parameters amongst others, as these are considered the hard constraints in practical deployments. A recent study showed that, across architectures, accuracy and inference time are in a hyperbolic relationship[6] so that small increments in accuracy cost large increases in inference time. One way to reduce the inference time when training a network is to reduce the input parameters into a network through smaller training images, which will lead to lower overall accuracy of the trained network (Fig. 1). However, as CNNs initially learn rough features and attend to the details later in the training process[14], we could start training with small images and lower information content and increase image size and information content towards the end. This should allow the network to reach a similar accuracy as if it was trained with large images from the start but at lower inference time (Fig. 1). In other words, overall training accuracy should be comparable regardless of the starting image size, but inference time should be lower when training is started with small images and, hence, training should be more efficient.

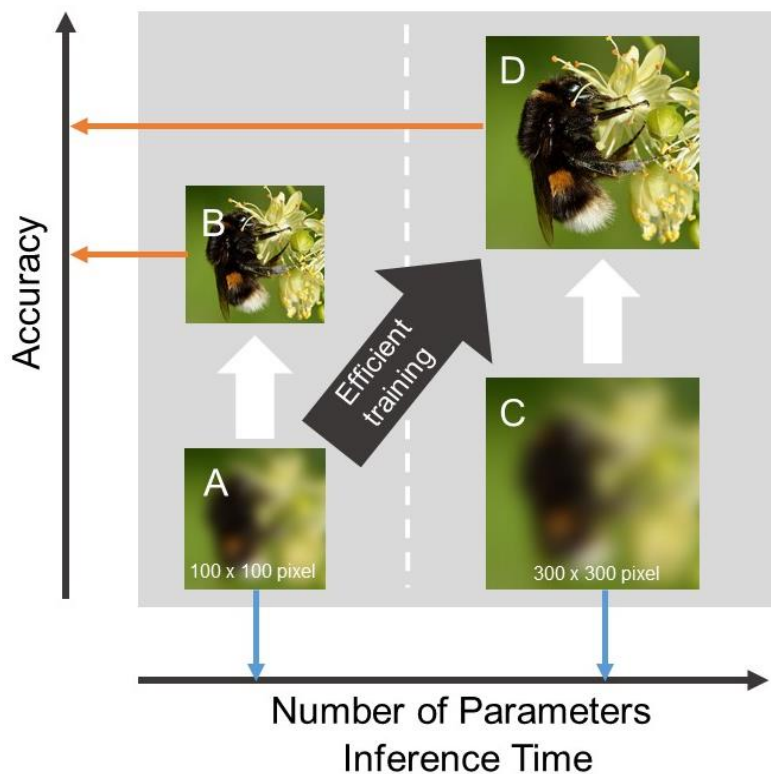

**Figure 1.** The 'Efficient Network Training (ENTR) Hypothesis' builds on the premise that parameter input determines inference time and network accuracy based on increasing information content of the network. If a network is usually trained with small images (A ➔ B), information content in the network is initially small and then increases. Maximum accuracy (orange arrows) and inference time (blue arrows) is lower compared to a network that is trained on large images (C ➔ D). However, to make the training process more efficient (A ➔ D), one can use the 'best of both world': start training with small images and increase to larger images later in the training process. This should be possible, because Convolutional Neural Networks initially learn rough (gradients and edges, indicated by blurred images) and later fine (details, indicated by sharp images) image aspects.

Here, we test the hypothesis that initially reducing image size during network training increases overall training efficiency (hereafter *'Efficient Network TRaining [ENTR] Hypothesis'*). We trained pre-trained Residual Network (ResNet) models (ResNet18, ResNet34, and ResNet50) on Google Colaboratory GPUs ('Colab')[15] with three small datasets from different scientific disciplines (material sciences, geosciences, and environmental sciences). We used three training regimes to test the ENTR hypothesis: 1) no initial decrease (control), 2) gradual increase (treatment 1) or 3) single-step increase in image size during training (treatment 2). To prove the ENTR hypothesis, an inference time-standardized accuracy measure of the treatments, but not necessarily accuracy *per se*, should improve relative to the control, across datasets and ResNet models. Our results support the ENTR hypothesis that initial image size decrease during training can increase training efficiency, which can be interpreted as a network regularization mechanism.

## Methods

*Datasets*
This study uses three small datasets (Tab. 1). The first is a new dataset of steel microstructures (hereafter *'Steel dataset'*), which was compiled from Google Image Search images. We downloaded images of iron-based steel alloys, which represent common forms of crystallized steel microstructures and form during material manufacturing and treatment[16]. Our dataset contains microscopic images of polished and etched surfaces of solid metallic materials. The identification of steel-based crystallized microstructures were subject to expert level verification, because the images can be very similar in some metallic phases depending on resolution, etching method, and quality (Fig. 2 A-D). In total, the dataset contains 1460 images in 7 classes.

The second is the UC Merced dataset (hereafter *'UCM dataset'*)[17] that is widely used in remote sensing image retrieval and scene classification as a benchmarking dataset[18]. The dataset can be downloaded here: http://weegee.vision.ucmerced.edu/datasets/landuse.html and contains 2100 images in 21 classes (Fig. 2 E-H).

The third is a new dataset of the common bee species of central Europe (hereafter *'Bees dataset'*)[19]. The identification of bees by field experts often depends on miniature features of the species and requires years of practice. Our final species list was cross-checked by a field expert and individual species combined into groups to allow for realistic species identification. We then downloaded images from Google Image Search based on the species list above. All pictures were manually verified and only included if they contained the bee in its natural environment (Fig. 2 I-L). In total, the dataset contains 3128 images in 22 classes.

**Table 1.** Dataset overview

| Dataset | Steel | UCM | Bees |
|---|---|---|---|
| *Image Category* | Microscopic | Aerial | Natural |
| *Within Image similarity* | High | Low | Intermediate |
| *Identification difficulty* | High | Low | Intermediate |
| *ImageNet resemblance* | None | Some | High |
| *Type* | New | Existing | New |
| *Total Images* | 1460 | 2100 | 3128 |
| *Classes* | 7 | 21 | 22 |

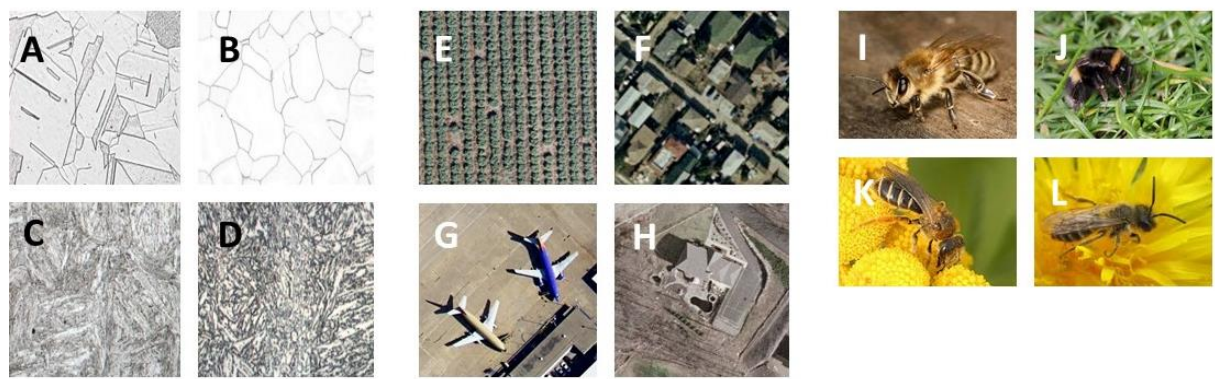

**Figure 2.** Sample images of the three datasets. Sample image classes are in the 'Steel dataset': A = Austenite; B = Ferrite; C = Martensite; D = Bainite; in the 'UCM dataset': E = Agriculture; F = Dense residential; G = Airplane; H = Medium residential; in the 'Bees dataset': I = *Apis mellifera*; J = *Bombus terrestris* group; K = *Lasioglossum calceatum*; L = *Andrena munituloides*

We chose these datasets for our analyses, because they differ in their representation in the ImageNet dataset that was used to pre-train our ResNet models and in their image types. The Bees dataset contains natural images that are very similar to the ones used in ImageNet, while the aerial images of the UCM dataset still represents much of the shapes and gradients that can be found in ImageNet[14] but not in the detailed images. The Steel dataset contains microscopic images that are not present in the ImageNet dataset. Consequently, we cover a range of datasets that a deep learning practitioner will encounter and, hence, assume that our findings show some generalization across common image recognition problems.

Maximum image resolution for all datasets was reduced to 250 pixels for best performance in Colab. All datasets were split randomly into training (90%) and validation set (10%).

*Model-Architectures*
For our experiment below, we compared ResNet18, ResNet34, and ResNet50 models. We chose these specific ResNet networks, because they have a favourable inference time per performed operation[6]. Moreover, they are deep neural nets (ResNet18, ResNet34 and ResNet50 contain 18, 34, and 50 layers, respectively), but do not suffer from other deep network problems such as vanishing gradients. We are not considering other deep networks such as DenseNets[20], which are not feasible for smaller datasets, take a very long time to run, and are difficult to train due to backpropagation problems.

*Experiment*
To understand the effect of initially reducing the size of training images on standardized accuracy (square root of accuracy/ inference time) for our three datasets and three ResNet models, we used three training regimes: i) 'no increase': the control for which we trained the network with the maximum image size; ii) 'gradual increase': we began with reduced image size and gradually increased image size throughout the training process; iii) 'stepwise increase': we started training the network with a reduced image size and increased training image size to the maximum for a couple of epochs before re-training the pre-trained layers. All models were trained on Colab, which allowed to use a Google GPU for a 12h time limit before a new session has to be restarted. For details on the training regimes see Fig. 3.

*Model training steps*
For all training regimes and networks above, we used weights pre-trained on the ImageNet dataset. To make the network more resistant to image noise[22], and to reduce overfitting, we used a dropout ranging from 0.4 to 0.6, which was consistent within datasets and networks. All networks within datasets were trained to the same amount of epochs to make results comparable, but not to full convergence in case of the Steel and Bees datasets due to technical and time limitations in Colab.

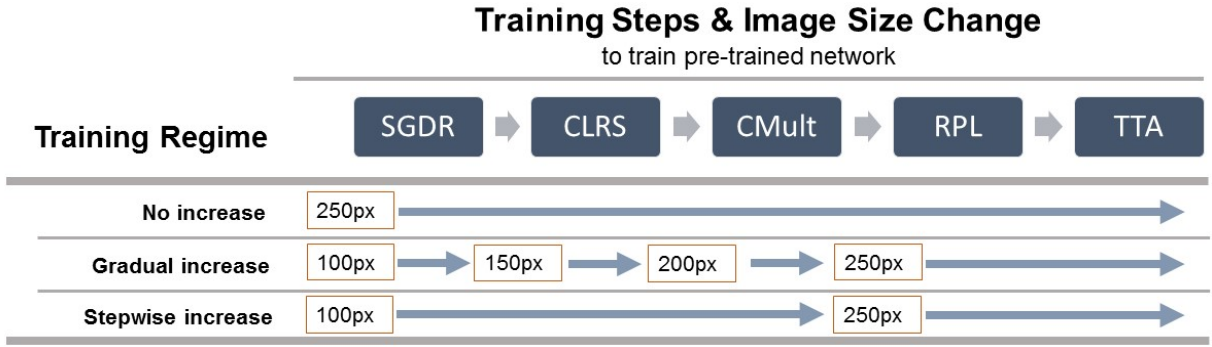

**Figure 3.** Schematic overview of the training steps and image size modifications in the training regimes. The following example illustrates the process: for the 'Gradual increase' training regime, we started with an image size of 100 pixels ('100px') before training with Stochastic gradient decent with restart ('SGDR'), and then gradually increased image size to 150, 200, and 250 pixels ('150px', '200px', '250px'), before training with the Cyclic learning rate scheduler ('CLRS'), Cyclic learning rate scheduler with multiple restarts ('CMult') and Re-training pre-trained layers ('RPL') and performing Test Time Augmentation ('TTA'), respectively. The training process is modified from [21].

For small datasets, convergence of the Stochastic Gradient Decent (SGD) can be a problem, because the number of gradient updates is limited (e.g., a dataset of 1,000 samples trained on 50 epochs will result only in 50,000 gradient updates). Consequently, the learning rate as the most important hyper parameter needs to be increased, out-of-sample-accuracy can fluctuate a lot, and convergence may not be achieved if training is stopped early. We used a learning rate finder to partially mitigate this problem by identifying a suitable learning rate before the start of the training session (for details see [23]).

We also used stochastic gradient decent with restart (SGDR), where the learning rate is automatically decreased with increasing proximity to the weight space's local minimum[23]. SGDR is opposed to common learning rate annealing, where a larger learning rate in the beginning of the training is manually decreased when the model stops improving. Moreover, we cyclically increased the learning rate to potentially jump between local minima in the weight space and find the resilient values. This process is also referred to as a 'cyclic learning rate scheduler'[24].

Noise, in particular in smaller datasets, is often compensated by image augmentation, i.e., geometric operations on the images to increase sample size. We used vertical flips, zooming and distortion of the sample images. However, augmentation can also be used to increase accuracy when all accuracy predictions of the augmented versions of an image are averaged. This approach is referred to as 'Test Time Augmentation' (TTA; see[25] and references herein). We included TTA after retraining the pre-trained layers.

Re-training pre-trained model layers (hereafter *'re-training'*) with decreasing learning rates per layer set can further improve model performance. This is based on the assumption, that lower level pre-trained layers to detect geometric features such as edges and gradients will not need much if any re-training; intermediate layers with more sophisticated features will take more training; and newly added top layers will require most training. This becomes evident when looking at the ImageNet visualization by Zeiler and Fergus[14]. We did not change re-training parameters within datasets and networks therein. For an overview of the model training steps see Fig. 3 and [21].

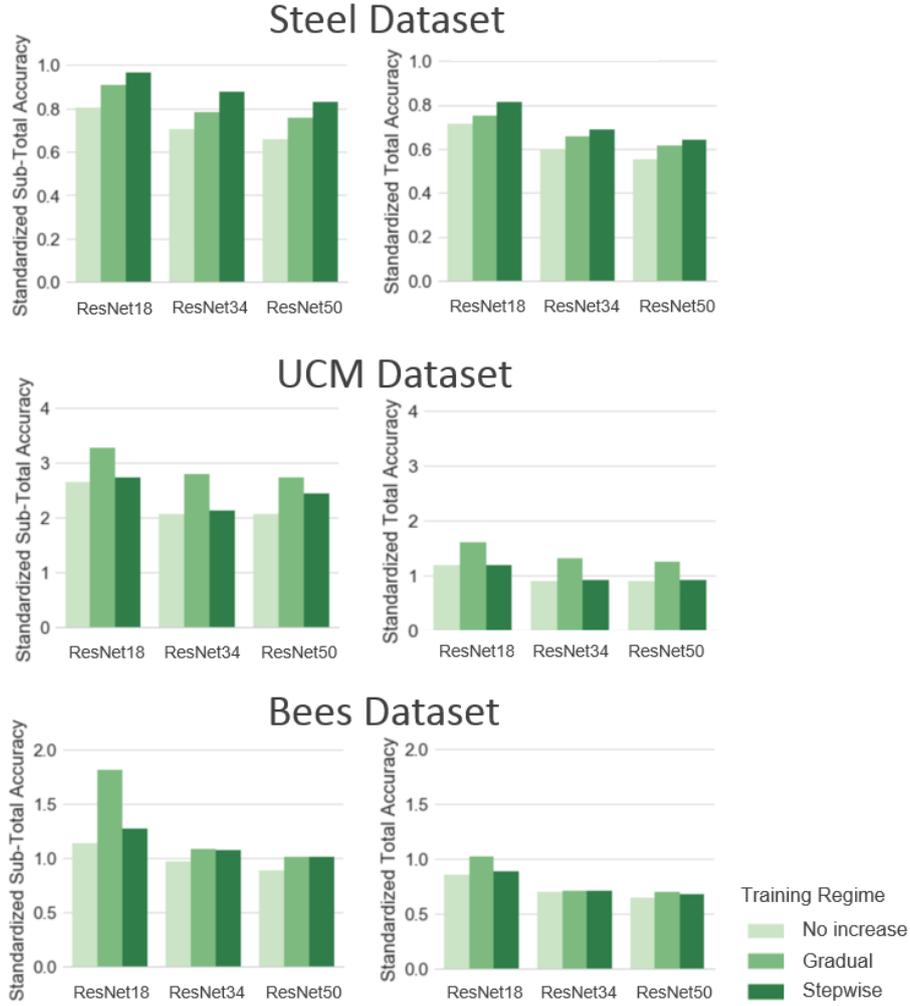

**Figure 4.** Experimental results for the network training across datasets and ResNet models measured as standardized accuracy. The first and second graph for each dataset shows standardized accuracy before ('Standardized Sub-Total Accuracy') and after ('Standardized Total Accuracy') re-training the pre-trained layers, respectively.

We separately compared standardized accuracy $As = \sqrt{A/T}$ (with $A$ being classification accuracy and $T$ being inference time) as a measure of efficiency and classification accuracy $A$ (hereafter *'accuracy'*) for each training regime per ResNet model and dataset before and after re-training. This is, because different representations of our datasets in the pre-training dataset may affect the outcome differently before and after re-training the networks.

**Results**

*Evaluating the 'Efficient Network TRaining' Hypothesis*
The standardized accuracy response to training regimes, where image size was increased was consistently higher across datasets and architectures. These results were also consistent before and after re-training. Specifically, for the most efficient training on the Steel dataset, our results show consistently that a stepwise image size increase yielded the best result with the ResNet18 network. In the UCM dataset, a gradual image size increase with the ResNet18 architecture was most efficient. The most efficient result for the Bee dataset was achieved with a gradual image size increase in the smallest ResNet18 network. Overall, this supports the ENTR Hypothesis that training image size increase leads to more efficient network training (Fig. 4).

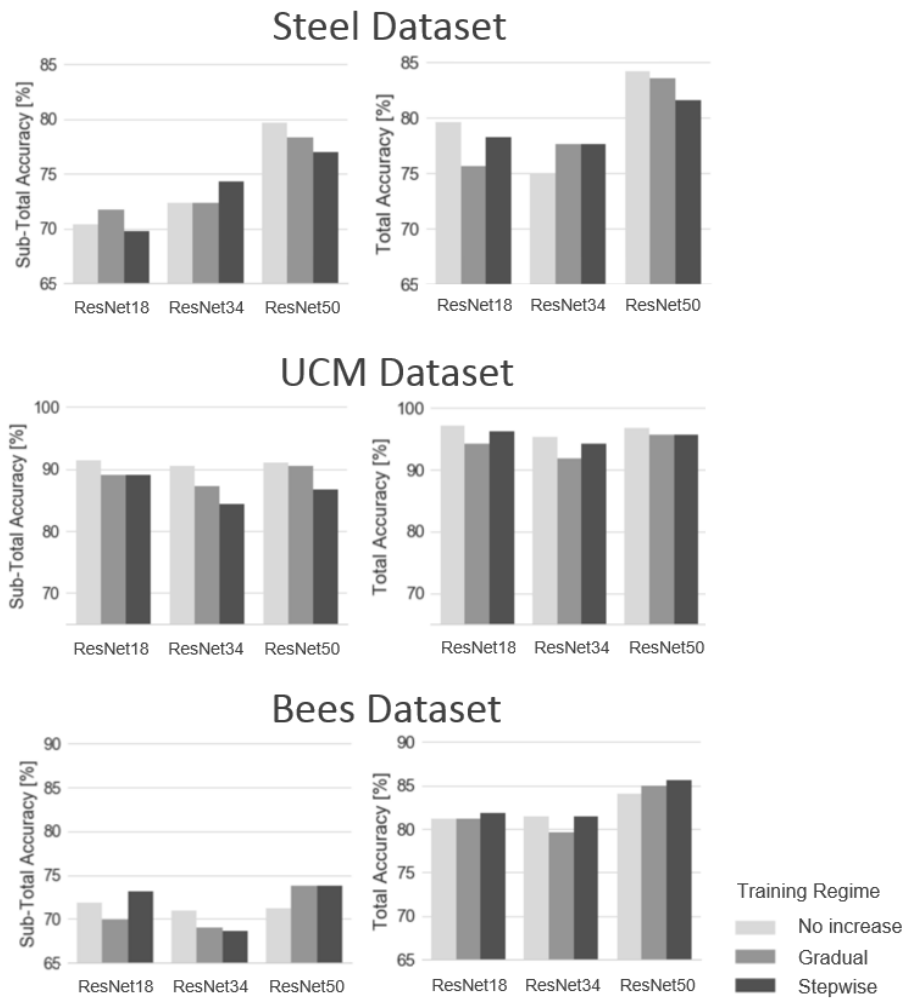

**Figure 5.** Experimental results for the network training across datasets and ResNet models measured as classification accuracy. The first and second graph for each dataset shows classification accuracy before ('Sub-Total Accuracy') and after ('Total Accuracy') re-training the pre-trained layers, respectively.

*Evaluating overall Accuracy Increase*
The accuracy response to training regimes was different across datasets and – except from the UCM dataset also within architectures. For all architectures in the UCM, most in the Steel and one in the Bees dataset, no image size increase led to the highest accuracy (Fig. 5). Specifically, the ResNet50 network trained on our Steel dataset yielded the highest accuracy with no image size modification regardless of pre or post accuracy of 79.6% and 84.2%, respectively. In the UCM dataset, the highest accuracies could be achieved with the ResNet18 network, which was consistent for pre (91.4% accuracy) and post (97.1% accuracy) training layer results. In the Bees dataset, a stepwise image size increase yielded the highest accuracy pre and post pre-training layers of respective 73.8% and 85.6% with the ResNet50 network.

**Discussion and Outlook**

In support of our 'Efficient Network TRaining (ENTR)' hypothesis, we show that initially decreasing the size of training images during the training process consistently improves training efficiency across datasets and ResNet models. This is an important contribution, because most network training improvements for image recognition tasks target accuracy but not efficiency improvements (see for example ImageNet, CIFAR, and Kaggle competitions). However, studies to find improvements in training efficiency are much needed to improve performance in practical applications[6].

Our results can be interpreted mechanistically as regularization. Network weights are initially pre-shaped on small images and subsequent large image training can then build on these pre-shaped weights. In networks trained with stochastic gradient decent (SGD), the sequence of learning images describes a trajectory across the parameter space, with a loss function converging when a local minimum is reached. Any perturbation early in the SGD training process increases the weights, which limits the regions of the parameter space that are accessible to the SGD. Later in the training process, larger parameter values mostly prevent shifting away from the specific parameter space[26]. Therefore, initially limiting and then increasing image size during network training may act as regularization that helps to generalize the network better and increases efficiency of model training.

While small datasets can pose a challenge for model convergence and final accuracy[8], they are well suited to work in constrained environments and can inspire research on standard datasets. We conducted our experiments on Google's free cloud computing GPU, Google Colab, which allows to do research with limited resources but comes with time constraints. In particular, the time limitations made it impossible to run model iterations until full convergence and several hundred replications to attribute solid errors to our estimates in Fig. 4 and Fig. 5. Despite this caveat, the consistent results across three different datasets and ResNet models suggest that initially decreasing image size in the training to make network training more efficient is more broadly applicable, potentially even to the standard ImageNet or CIFAR datasets.

There are a number of exciting future research avenues opening up. At first, we tested the ENTR hypothesis on new and already available small datasets to make our results robust and comparable. However, it will be interesting to see if the ENTR hypothesis holds not only for other small datasets, but also when it is tested on large and standard datasets such as ImageNet and CIFAR. Second, we proposed a mechanism (Fig. 1) that builds on parameter and inference time reduction, accuracy increase, and suitable information maximization that is then discussed in the regularization theory framework. It will be interesting to test the specific mechanisms behind the ENTR hypothesis and potentially come up with alternative explanations that could also be based in information theory. And lastly, it will be interesting to compare these mechanistic findings to other learning systems. For example, in structural learning in animals and humans, initial optimization of parameters helps to lay the structural foundation to address new tasks. The optimized parameters are then 'left' in a fixed and non-linear relationship so that the same structure can be used to address similar learning problems[27]. Thus, the ENTR Hypothesis could opened an avenue for exiting new research to improve and understand network training efficiency.

**Acknowledgements**
We thank Christian Bauckhage and Raphael Schween for very helpful discussions on the manuscript. Annika Haas checked the species list to compile the Bees dataset.